\documentclass{article}

\usepackage[margin=1in]{geometry}

\usepackage{graphicx}
\usepackage{url}            
\usepackage{booktabs}       
\usepackage{amsfonts}       
\usepackage{nicefrac}       
\usepackage{amsmath}
\usepackage{amssymb}
\usepackage{float}
\usepackage{lipsum}
\usepackage{booktabs} 
\usepackage{subcaption}
\usepackage{xcolor}
\usepackage{latexsym}
\usepackage{enumitem}
\usepackage{tablefootnote}
\usepackage[round,semicolon]{natbib}
\usepackage{xspace}
\usepackage{textcomp}
\usepackage{makecell}
\usepackage{multirow}
\usepackage{lscape} 
\usepackage{siunitx}
\usepackage{microtype}
\usepackage{url}            
\usepackage{booktabs}       
\usepackage{amsfonts}       
\usepackage{nicefrac}       
\usepackage{changepage}
\usepackage{xargs}          
\usepackage{wrapfig,lipsum,booktabs}
\usepackage{enumitem}
\usepackage{longtable}
\usepackage{makecell}
\usepackage{graphicx}
\usepackage{subcaption}
\usepackage[T1]{fontenc}
\usepackage{tgpagella}
\usepackage{xspace}  

\definecolor{mydarkblue}{rgb}{0,0.08,0.45}
\usepackage[colorlinks,citecolor=mydarkblue,urlcolor=mydarkblue,linkcolor=mydarkblue]{hyperref}

\usepackage[most]{tcolorbox}
\usepackage{bm}
\definecolor{myRed}{rgb}{1,0.0,0}

\definecolor{myGreen}{rgb}{0.0,1.0,0}

\definecolor{myBlue}{rgb}{0,0.0,1.0}

\definecolor{frenchblue}{rgb}{0.0, 0.45, 0.73}
\newcommand{\frenchblue}[1]{{\color{frenchblue}{#1}}}
\newcommand{\bluegain}[1]{\textbf{\frenchblue{(+#1\%)}}}

\definecolor{frenchlilac}{rgb}{0.53, 0.38, 0.56}
\newcommand{\frenchlilac}[1]{{\color{frenchlilac}{#1}}}
\newcommand{\purplelost}[1]{\textbf{\frenchlilac{(-#1\%)}}}

\title{
\vspace{-4em}%
  \vspace{0.1em}%
  \center
  \vskip 0.4in%
  \vskip -\parskip%
 \fontsize{18}{18}{\textbf{Transcending Scaling Laws with 0.1\% Extra Compute
}}
  \vskip -\parskip%
  \vskip 0.01in}

\author{
\normalsize{}
 \textbf{
Yi Tay \hspace{3mm} Jason Wei \hspace{3mm} Hyung Won Chung \hspace{3mm} Vinh Q. Tran  \hspace{3mm} David R. So \hspace{3mm} Siamak Shakeri} \vspace{1mm} \\
\normalsize{}{
\textbf{Xavier Garcia \hspace{3mm} Huaixiu Steven Zheng \hspace{3mm} Jinfeng Rao \hspace{3mm}  Aakanksha Chowdhery \hspace{3mm} }}  \vspace{1mm} \\
\normalsize{}{
\textbf{Denny Zhou \hspace{3mm} Donald Metzler \hspace{3mm} Slav Petrov \hspace{3mm}  Neil Houlsby \hspace{3mm}}} \vspace{1mm} \\
\normalsize{}{
\textbf{Quoc V. Le \hspace{2mm} Mostafa Dehghani}} \vspace{2mm} \\ 

\\
Google
 }

\newcommand{\methodname}{UL2R\xspace}
\newcommand{\modelname}{U-PaLM\xspace}
\date{}
\begin{document}

\maketitle

\sloppy

\begin{abstract}
\noindent Scaling language models improves performance but comes with significant computational costs.
This paper proposes \methodname, a method that substantially improves existing language models and their scaling curves with a relatively tiny amount of extra compute. The key idea is to continue training a state-of-the-art large language model (e.g., PaLM) on a few more steps with UL2's mixture-of-denoiser objective. We show that, with almost negligible extra computational costs and no new sources of data, we are able to substantially improve the scaling properties of large language models on downstream metrics. In this paper, we continue training PaLM with \methodname, introducing a new set of models at 8B, 62B, and 540B scale which we call \modelname. Impressively, at 540B scale, we show an approximately 2x computational savings rate where \modelname achieves the same performance as the final PaLM 540B model at around half its computational budget (i.e., saving $\sim$4.4 million TPUv4 hours). 

\vspace{3mm}
\noindent We further show that this improved scaling curve leads to ``emergent abilities'' on challenging BIG-Bench tasks---for instance, \modelname does much better than PaLM on some tasks or demonstrates better quality at much smaller scale (62B as opposed to 540B). Overall, we show that \modelname outperforms PaLM on many few-shot setups, i.e., English NLP tasks (e.g., commonsense reasoning, question answering), reasoning tasks with chain-of-thought (e.g., GSM8K), multilingual tasks (MGSM, TydiQA), MMLU and challenging BIG-Bench tasks. Finally, we provide qualitative examples showing the new capabilities of \modelname for single and multi-span infilling.

\end{abstract}

\begin{figure}[H]
     \centering
     \begin{subfigure}[b]{0.5\textwidth}
         \centering
         \includegraphics[width=\textwidth]{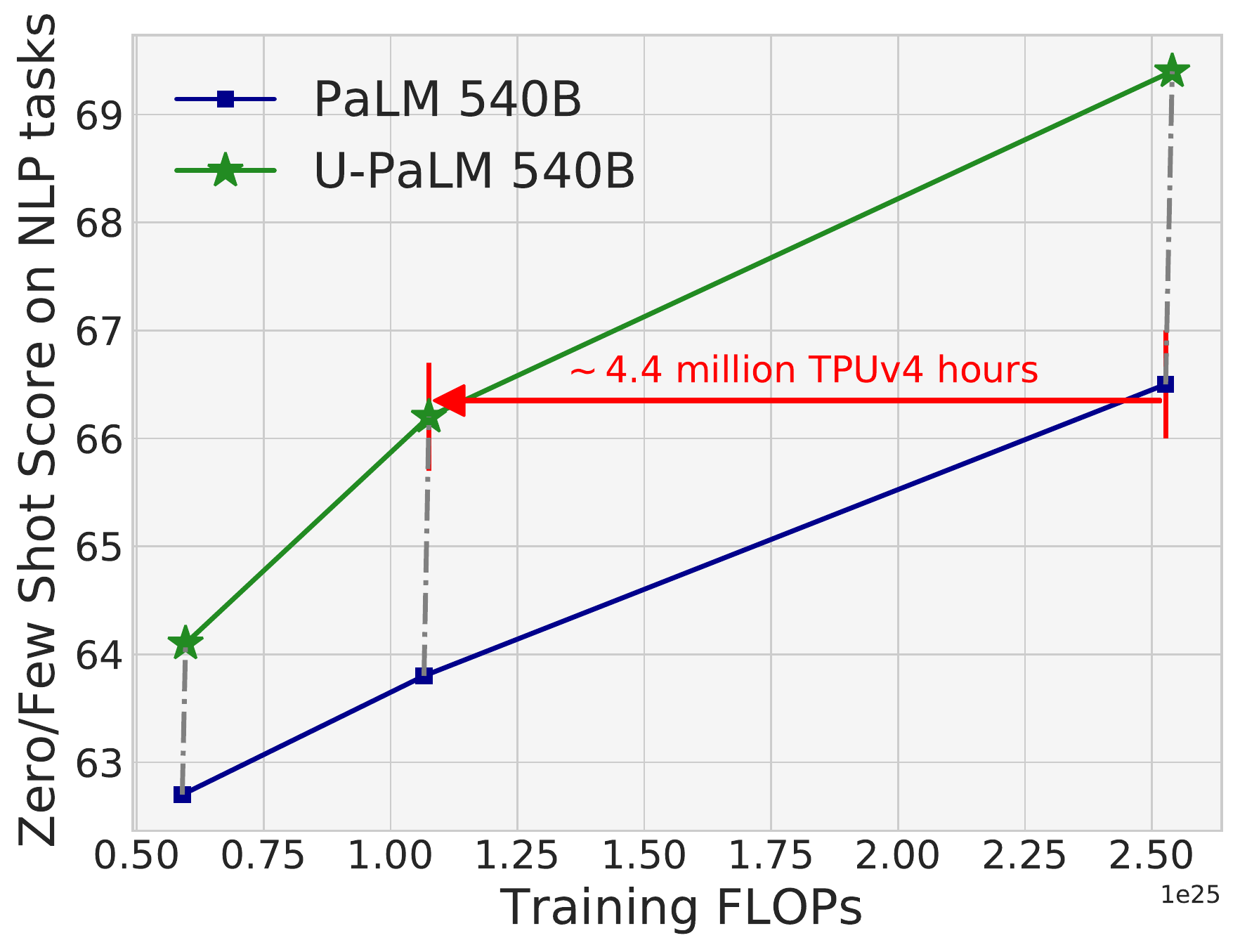}
     \end{subfigure}
    \caption{Compute (training flops) versus Quality (average of 20+ NLP zero and few-shot tasks listed in Appendix~\ref{app:fewshot-exp-details}). The black dotted line shows the path from initialization from a PaLM checkpoint and training further with \methodname.}
    \label{fig:teaser}
\end{figure}

\newpage

\section{Introduction}

There has been significant interest in scaling of language models~\citep{rae2021scaling,chowdhery2022palm,brown2020language}. Scaling has inspired new research across multiple fronts, e.g., scaling laws~\citep{kaplan2020scaling,hoffmann2022training,tay2022scaling}, emergent abilities~\citep{wei2022emergent,ganguli2022predictability}, reasoning capabilities~\citep{wei2022chain,lewkowycz2022solving}, inter alia. Generally, scaling laws predict a continued improvement in language model quality as we continue to scale up the computational budget (e.g., bigger models or more data). To date, most large language models that form the basis of scaling law research are trained almost exclusively as left-to-right causal language models~\citep{kaplan2020scaling,hoffmann2022training}.

This paper proposes a new method to dramatically improve the scaling curves of large language models on downstream performance with a relatively tiny amount of additional computation cost. The key idea is to continue training an existing causal language model~\citep{chowdhery2022palm} with a mixture of new objectives---specifically, the UL2 training objective mixture~\citep{tay2022unifying}. This restoration is expected to only cost roughly $0.1\%$ to $1\%$ of the original training FLOPs and requires no new data sources, making it highly efficient and convenient. We call this approach \methodname or UL2Restore.

The UL2 objective combines prefix language modeling and long-short span corruption (e.g., infilling) tasks~\citep{raffel2019exploring} that can be controlled at inference time using a mode switching prompt. Training a large language model with UL2 can be interpreted as teaching it to leverage bidirectional attention (i.e., PrefixLM) or leverage infilling-style pretraining that have been the foundation of language understanding (e.g., T5~\citep{raffel2019exploring}). To this end, we postulate that imbuing a state-of-the-art large language model such as PaLM~\citep{chowdhery2022palm} with these diverse pretraining schemes as a complement to the original language model objective, enables the model to perform significantly better. Moreover, the UL2 objective enables new prompting capabilities in PaLM which allows it to perform infilling based prompting. 

We show that adapting PaLM with \methodname{} not only results in significantly better scaling laws on well-established few-shot NLP tasks, but also, in our scaling experiments on downstream few-shot tasks, we show that \methodname is two times more efficient (computation savings of approximately 2x) at 540B scale - reaching the performance of the final PaLM 540B model with only half the computation, saving up to 4.4 million TPUv4 hours. 

In addition to competitive performance across a range of well-established NLP~\citep{wang2019superglue}, multilingual~\citep{clark2020tydi,shi2022language}, and reasoning~\citep{cobbe2021training} benchmarks, we also study the impact of \methodname on a suite of challenging BigBench tasks from \citet{wei2022emergent}. Notably, a subset of tasks are described as `emergent` because PaLM's performance remains flat up to model scale of 62B and only becomes better than non-random at 540B scale. On these set of tasks, we find that \methodname enables (1) doing significantly better at tasks that PaLM struggles at (e.g., navigate, geometric shapes, hyperbaton) and (2) elicits emergent behavior at a smaller scale such as 62B or 8B (e.g., crass ai, vitaminc fact verification). On top of that, \modelname strongly outperforms PaLM on some challenging BigBench tasks.

Emergence within the context of large language models is a nascent research area. As the Nobel prize-winning physicist Philip Anderson put it, \textit{`More is different.`}~\citep{anderson1972more} which describes unpredictable phenomena at different scales. In our context and with mixture-of-denoisers in UL2, we would like to think of this phenomena as \textit{`More is different, but different can also more'} since different pretraining objectives can improve language model quality or elicit new emergent abilities. This work shows that diversity and richer training paradigms can be key to learning new capabilities that were previously hard to acquire with only causal language modeling.

Finally, in addition to emergent task performance and overall improved scaling curves, we show that \modelname is also practically more useful since it is equipped with a secondary mode of prompting, i.e., bidirectional infilling. Specifically, \methodname enables a secondary capability for prompting \modelname which can be used to fill in more than one blanks in the input prompt. Interestingly, we find that only a small amount of \methodname (e.g., 0.1\% tokens or FLOPs) is sufficient to imbue the model with this new capability.

\section{Related Work}
\paragraph{Large language models} Scaling and improving large language models is one of the most impactful research areas in modern artificial intelligence~\citep{chowdhery2022palm}. To this end, large language models not only continue to improve as we scale in terms of data or computational budget~\citep{hoffmann2022training,kaplan2020scaling} but also acquire new abilities~\citep{wei2022emergent}. The impact of large language models has been ubiquitous and pervasive, unlocking breakthroughs across many fields, e.g., reasoning~\citep{wei2022chain, wang2022self, zhou2022least, drozdov2022compositional}, math~\citep{lewkowycz2022solving}, dialog~\citep{thoppilan2022lamda}, multimodal applications~\citep{yu2022scaling}, retrieval~\citep{tay2022transformer} \textit{inter alia}.

While there have been many paradigms and self-supervision methods proposed to train these models~\citep{devlin2018bert,clark2020electra,yang2019xlnet,raffel2019exploring}, to this date most large language models (i.e., more than 100B parameters) are trained as decoder-only casual language models. For example, flagship large language models such as GPT-3~\citep{brown2020language}, Gopher~\citep{rae2021scaling} and PaLM~\citep{chowdhery2022palm} are all trained as causal language models. Meanwhile, bidirectional models (e.g., BERT~\citep{devlin2018bert}, T5~\citep{raffel2019exploring}, ST-MoE~\citep{stmoe}) have also been very popular as the goto model of choice, especially in smaller computational regimes (e.g., less than 30B parameters and often times in the ranges of hundred of millions of parameters).

\paragraph{Scaling laws of large language models} \citet{kaplan2020scaling} investigated scaling laws of Transformer language models and first showed the scaling laws are predictive of future performance. The authors found that model size (and not shape) correlates strongly with model quality, i.e., upstream cross entropy. \citet{tay2021scale} studied the scaling properties of encoder-decoder models and their impact on upstream and downstream finetuning tasks. Generally, \citet{tay2021scale} found that upstream perplexity and downstream quality does not always correlate. As a follow up, \citet{tay2022scaling} studied the scaling laws of different model architectures and found that inductive bias does significantly impact the scaling behavior of the model. Finally, \citet{hoffmann2022training} proposed compute-optimal models that popularized the \textit{`chinchilla'} scaling laws - an approach that aims to be predictive of the optimal amount of data given the number of model parameters. In this work, we mainly consider scaling laws over downstream performance largely because this is more reflective of a language model's usability. Since downstream performance is more important than upstream cross entropy, we advocate for future scaling studies to always incorporate downstream evaluation (and metrics) as opposed to only using cross entropy loss.  

\paragraph{Emergent Abilities}
New behaviors that arise due to scaling language models have been increasingly referred to as \textit{emergent abilities}~\citep{jacobsdefinition,ganguli2022predictability,wei2022emergent}.
For instance, \citet{wei2022emergent} define emergent abilities as ``abilities that are not present in smaller models but as present in larger models.''
For a few-shot prompted task, this would look like a flat scaling curve (random performance) until a certain critical threshold, during which performance increases to substantially above random.
This type of phenomena has been observed across dozens of tasks in the BIG-Bench benchmark~\citep{srivastava2022beyond}.
Although such emergent abilities are typically observed as a function of scale, increasing model scale to induce emergent abilities is computationally expensive.
In this paper we show how \methodname unlocks emergence without increasing the number of model parameters.



\paragraph{Continued Training of Language Models} 
The paradigm of continue to train (or finetune) a language model on more data or tasks is commonly known as adaptation. 
A range of prior work has shown that finetuning language models on a collection of NLP tasks can improve downstream performance on a broad range of downstream tasks~\citep[][\textit{inter alia}]{aghajanyan2021muppet,aribandi2022ext5,wei2021finetuned,sanh2022multitask,ouyang2022training}.
The majority of this prior work, however, requires additional data such as aggregating dozens or hundreds of NLP datasets~\citep{raffel2019exploring,aghajanyan2021muppet,aribandi2022ext5}, writing additional templates of instructions~\citep{wei2021finetuned,sanh2022multitask}, or finetuning on human-labeled annotations~\citep{ouyang2022training}.
\methodname does not require new data since it simply re-uses the pre-training data, which makes it orthogonal to continued training methods that leverage large collections of NLP datasets.
Adapting a pretrained language model with a new self-supervised objective has been explored. For example, a model trained with a language modeling objective can be adapted by further training with the masked language modeling objective~\citep{wang2022language}. The other direction is also possible; a model trained with a masked language objective can be adapted with the causal language modeling objective~\citep{wang2022language,lester2021power}.
\methodname follows a similar idea but uptrains a language model with a set of diverse and new preordaining tasks from mixture-of-denoisers, even after a vast amounts of standard pretraining and demonstrates a very rapid improvement on variety of setups and tasks.

\paragraph{Unified language learner (UL2) } The UL2~\citep{tay2022unifying} model is a state-of-the-art model that bridges both generative causal language models and bidirectional language models. UL2 proposes a mixture-of-denoiser objective that mixes prefix (non-causal) language modeling and infilling (span corruption) within the same model and leverages \textit{mode prompts} to switch between modes during downstream tasks. UL2 is architecture agnostic in which the authors argue that the choice of decoder-only versus encoder-decoder models is largely an efficiency trade-off. In~\citep{tay2022unifying}, the final UL2 model was trained as a 20B encoder-decoder model, which achieves very compelling performance on both finetuning and in-context learning.


\section{\modelname}
This section introduces the technical details of \modelname (i.e., PaLM + \methodname). \modelname is initialized from PaLM and leverages the same architecture. This section describes the training procedures of \methodname and how they are applied to continue training PaLM.
\subsection{Training Data} To keep things consistent, we train this model with the same data mixture as PaLM and do not rely on additional sources of data (labeled or unlabeled). 

There are three main reasons for this choice. Firstly, we did not want to introduce new tokens to our training process which could conflate findings. Secondly, we did not want to over-index on scaling studies that only measure impact on upstream cross entropy~\citep{hernandez2022scaling} which claims that repeating data in small quantities could be dis-proportionally harmful. Since the empirical results we obtained are strong, we postulate that repeating tokens could perhaps be not harmful at smaller quantities after all. This is also backed by the continued training of PaLM 62B in~\citep{chowdhery2022palm} which showed that repeated data could result in small gains, albeit not as strong as fresh tokens. Thirdly, we consider our data transformation (via UL2) on the training data sufficiently unique and therefore prevents us from explicitly training on the same data with the exact objective or suffering from any memorization issues. 
\subsection{Prefix Language Model Architecture}
We train \modelname using the prefix language model (PrefixLM) architecture, also sometimes known as a non-causal decoder-only model. The PrefixLM architecture keeps a non-causal mask in its prefix (or inputs) and applies bidirectional attention to input tokens. 

In this architecture, we use a total combined sequence length of $2048$ (e.g., PaLM's sequence length) which is then split to 1024 inputs and 1024 targets. In the original UL2 paper and infrastructure, an artifact of its preprocessing pipeline applies padding tokens \textit{first} before combining \texttt{inputs} and \texttt{targets}. For decoder-only language models, this is inefficient since we would end up with a concatenation of \texttt{[prefix] [prefix's padding] [target]}. 

In this work, we optimize the Prefix padding by forcing the model to concatenate prefix and target \textit{before} applying any additional padding. Packing, trimming and padding is then subsequently applied later after the prefix has been concatenated with the targets. Through this \textit{prefix optimization}, we are able to improve example-level sample efficiency of the model.

\subsection{Loss Objectives}
\label{sec:objectives}
This section describes the setting for the UL2 mixture-of-denoisers that we use in \methodname. The UL2 mixture-of-denoiser objective comprises of three types of denoisers.
\begin{itemize}
    \item \textbf{Regular denoising} whereby the noise is sampled as spans, replaced with sentinel tokens. This is also the standard span corruption task used in \citet{raffel2019exploring}. Spans are typically uniformly sampled with a mean of $3$ and a corruption rate of $15\%$.
    \item \textbf{Extreme denoising} whereby the noise is increased to relatively \textit{`extreme`} amounts in either a huge percentage of the original text or being very long in nature. Spans are typically uniformly sampled with a mean length of $32$ \textbf{OR} a corruption rate of up to $50\%$.
    \item \textbf{Sequential denoising} whereby the noise is always sampled from the start of the text to a randomly sampled point in the text. This is also known as the PrefixLM objective (not to be confused with the architecture).

\end{itemize}

We kept this simple since many ablations were already explored in \citet{tay2022unifying}. We kept the original 7 denoisers as the initial version but later found that a mixture of only three tasks, e.g., $50\%$ PrefixLM, 25\% Long (extreme) span corruption, and 25\% regular span corruption to be quite simple and efficient for the setup of continued training. We kept the original mode prompting tokens in the original UL2 design. We used \texttt{[S2S]} for S-denoisers (PrefixLM), \texttt{[NLU]} for R-denosiers and \texttt{[NLG]} for X-denoisers. The 540B \modelname model was mainly trained with 50\% S-denoiser (PrefixLM), 25\% R-denoisers, and 25\% X-denoisers.

\subsection{Training}
We train the 540B model for a total of 20k steps with a batch size of $32$. We mildly ablate these settings in early experiments with 62B and 8B models but keep them capped within a certain ballpark (e.g., 128 batch size for 50k steps). As a result, this is more similar to \textit{`finetuning'} as compared to full pretraining. The number of additional tokens is therefore very negligible compared to the original pretraining run often coming in at around or less than $0.1\%$ additional compute. The total number of extra tokens we train on for the 540B model is approximately 1.3 Billion which constitutes 0.16\% extra computation. We use a cosine learning rate decay schedule that anneals the learning rate from $10^{-4}$ to $10^{-6}$. Notably, we  also tried a low constant learning rate and found them to perform quite identically. Our \modelname 8B and 62B models are trained using 64 TPUv4 chips. Training an \modelname 540B model only consumes 512 TPUv4 chips and finishes in about 5 days which is considered to be lightweight.


\section{Experiments}
This section reports the experimental results of \modelname.

\subsection{Improved Scaling Properties on Few-shot Learning}
In this experiment, we show improved scaling curves from small amounts of \methodname training on top of both PaLM 8B and PaLM 540B. We use downstream metrics and few-shot evaluation since (1) this is closer to usability of these models and (2) loss with UL2 and causal language modeling is not comparable. We initialized and trained multiple \modelname models using different PaLM intermediate checkpoints. On the 8B model, we repeated this 7 times at different intervals. Given that the 540B model was more computationally demanding, we only managed to fit 3 points. For evaluation, we use the average score of NLU and NLG tasks from the GPT-3 suite~\citep{brown2020language}. In total we use 26 tasks (e.g., TriviaQA, NaturalQuestions, SuperGLUE, PIQA, OpenbookQA, ANLI etc). Details and exact scores for Figure \ref{fig:scaling} can be found in the Appendix. 

\begin{figure}
     \centering
     \begin{subfigure}[b]{0.45\textwidth}
         \centering
         \includegraphics[width=\textwidth]{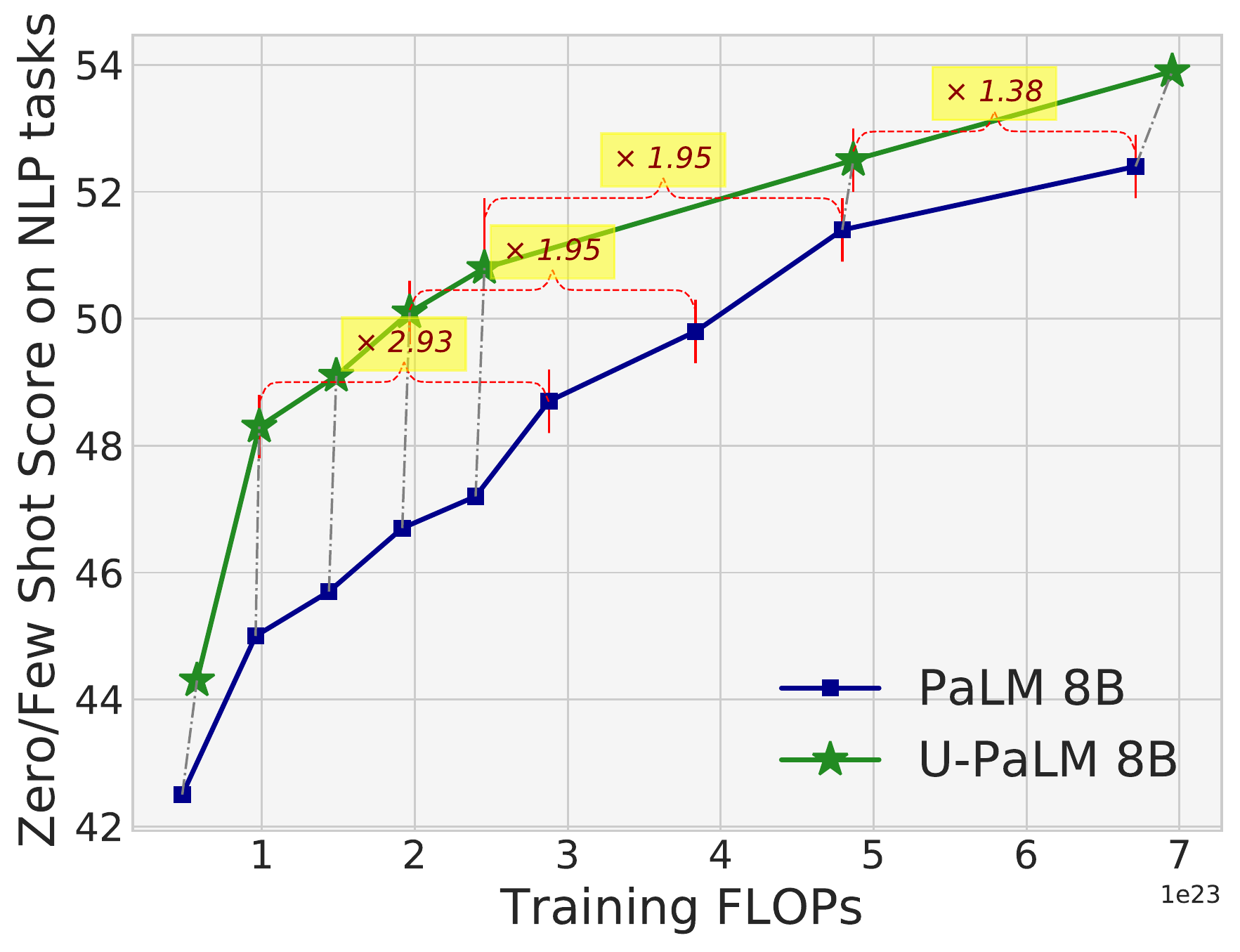}
        \label{fig:8b_scaling}
     \end{subfigure}
     \hspace{5pt}
     \begin{subfigure}[b]{0.45\textwidth}
         \centering
         \includegraphics[width=\textwidth]{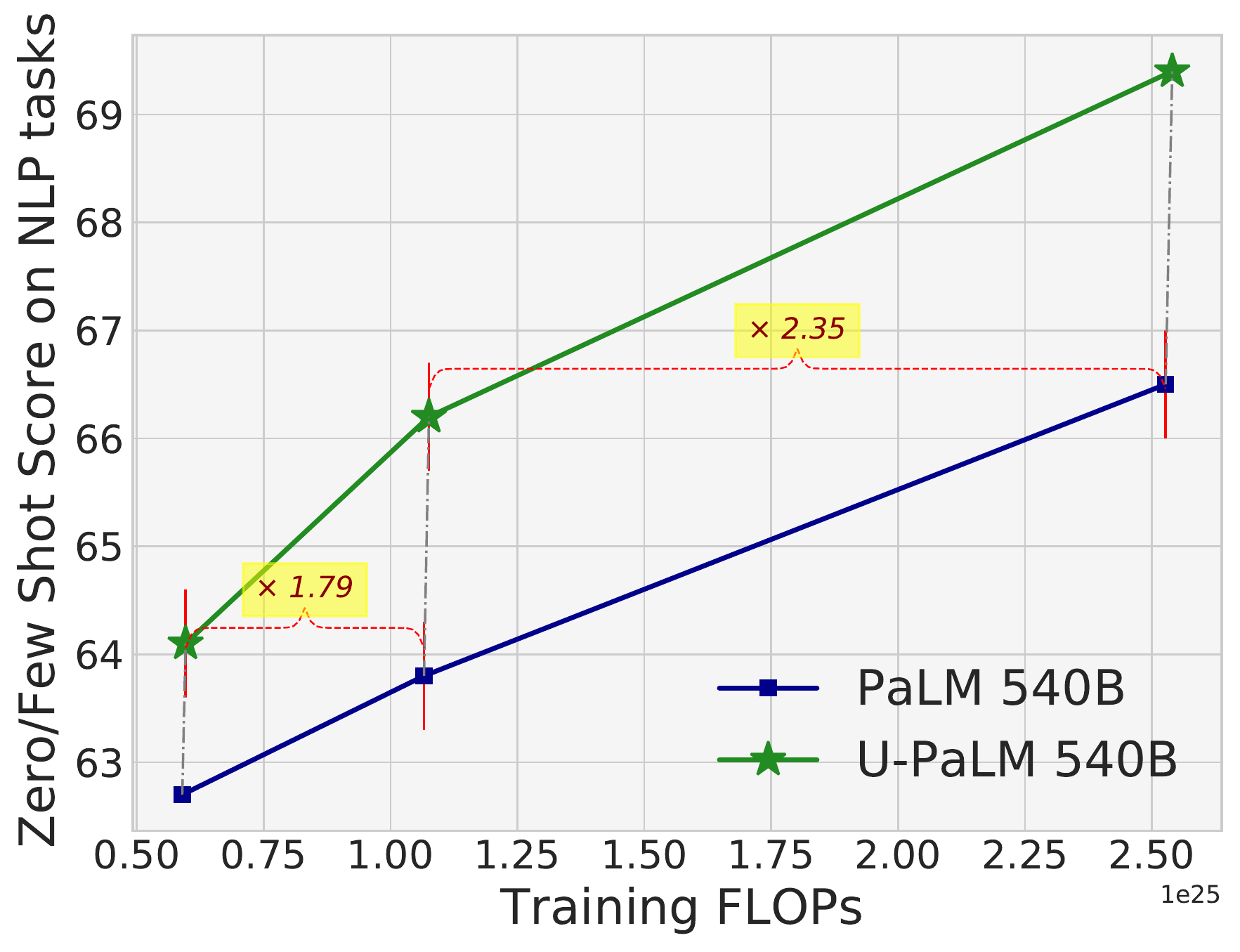}
        \label{fig:540b_scaling}
     \end{subfigure}
    \caption{Computation cost (training flops)~\citep{dehghani2021efficiency} versus Quality (average of 20+ NLP zero and few-shot tasks). The dotted line shows the path from initialization from a PaLM checkpoint and training further with \methodname. These plots also present pairs of PaLM and \modelname models with comparable/similar performance along with the ratio of PaLM computation cost vs the corresponding \modelname computation cost.  For example, PaLM 540B trained for $\sim2500$ zFLOPs (right most point) took $\sim2.35$ times of the computation cost of \modelname 540B trained for $\sim1075$ zFLOPs, while both models are comparable in terms of performance on zero/few shot on NLP tasks.}
    \label{fig:scaling}
\end{figure}

Figure \ref{fig:scaling} shows that \modelname substantially outperforms the original PaLM models both at 8B scale and 540B scale. Note that the dotted lines represent a pathway before and after \methodname training, we show that \methodname training improves the scaling curve of PaLM substantially, i.e., \methodname provides a more compute-efficient performance improvement compared to training the original PaLM models for longer with the standard causal language modeling objective.

\paragraph{8B versus 540B} Generally, \methodname consistently improves the underlying PaLM models. Nevertheless, we observe different behaviors on the 8B and 540B models. The gap seems to narrow as the performance of PaLM 8B starts to plateau, i.e., the largest gains are near to the middle of training. As for 540B, the gain continues to grow even at 780B tokens. We believe that this is due to the fact that PaLM 540B still has significant headroom beyond 780B tokens.

\paragraph{Savings Rate} At a certain stage of training, we have an option to continue training for K more steps using the standard causal language modeling objective OR applying \methodname for a small amount of steps. Here we discuss the counterfactual savings rate of choosing \methodname as opposed to continue training with caussal language modeling. For the 540B model, the saving rates at the middle checkpoint is approximately 2x. This is equivalent to about 4.4 million TPUv4 hours for the 540B model. For the 8B model, the saving rate tend to be lowest at both the start and convergence of the model. It seems to be higher at middle stages of training (relative to convergence) which shows that the utility of \methodname changes with respect to the amount of causal language modeling training already done. For the 540B model, since the PaLM model was not trained to convergence and the number of tokens to parameters ratio is relatively low, the savings rate could still be increasing even beyond 2.35x. Overall, the amount of savings is quite proportionate to the point of training and stage of convergence of the model and can probably be predicted by standard scaling laws~\citep{kaplan2020scaling,hoffmann2022training}.

\begin{figure}[H]
     \centering
     \includegraphics[width=\textwidth]{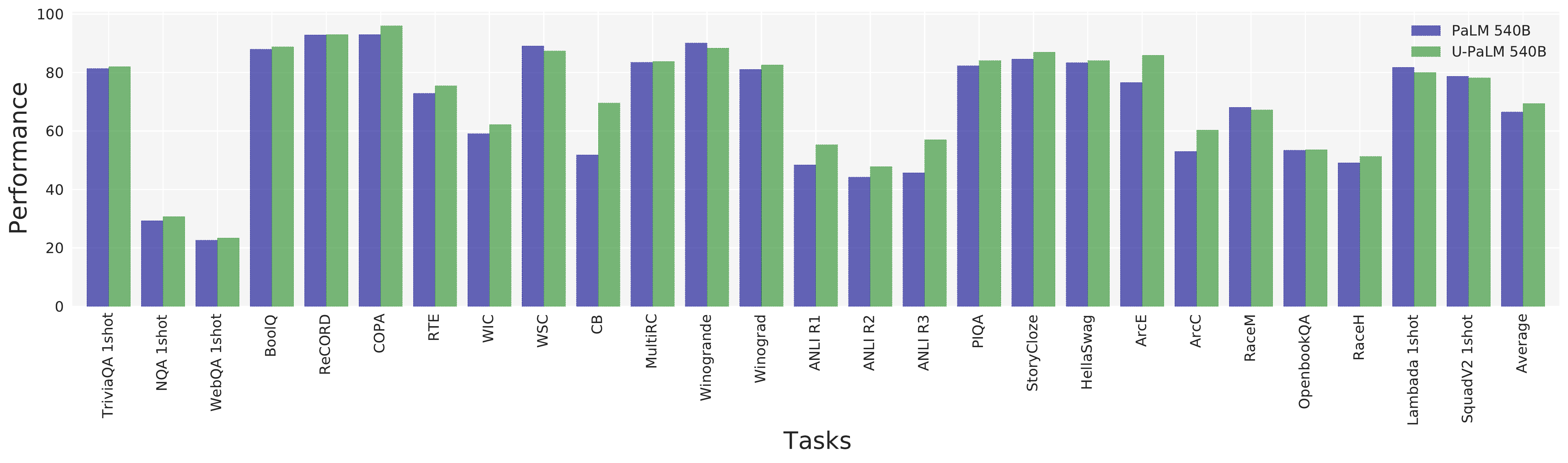}
    \caption{Break down scores of individual zero-shot and one-shot NLP tasks for PaLM and \modelname 540B trained for 780B tokens. \modelname outperforms PaLM 540B and achieves SOTA on 21 out of 26 tasks.}
    \label{fig:fewshot_nlp_tasks}
\end{figure}

\paragraph{Breakdown on individual tasks} FIgure \ref{fig:fewshot_nlp_tasks} reports the individual scores on each zero and one-shot task in the mixture. We show that \modelname 540B outperforms PaLM 540B on 21 out of 26 tasks. Given that PaLM is the SOTA language model on these tasks, this makes \modelname the new state-of-the-art on these tasks.

\subsection{BigBench Emergent Suite} 

\begin{table}[H]
    \centering
    \small
    \begin{tabular}{lp{4.5cm}cl}
    \toprule
    task  & task /reasoning type& PaLM 540B & \modelname 540B \\ 
    \midrule
    \texttt{navigate} & arithmetic, logical & 55.3 & \textbf{67.0} \bluegain{21.2} \\
    \texttt{strategyqa}  & multi-step  & 73.9 & \textbf{78.3} \bluegain{6.0} \\
     \texttt{crass\_ai} & commonsense  & 97.7 & \textbf{100} \bluegain{2.4} \\
      \texttt{logical\_sequence} & commonsense  & \textbf{92.3} & 86.5 \purplelost{6.7} \\
    \texttt{vitaminc\_fact\_verification}  & contextual, commonsense & 70.2 & \textbf{73.9} \bluegain{5.3} \\
    \texttt{understanding\_fables}& commonsense &75.7 & \textbf{78.4} \bluegain{3.6} \\
    \texttt{identify\_odd\_metaphor} & analogical& 87.2 & \textbf{87.5} \bluegain{0.3}\\
      \texttt{hyperbaton} & contextual QA & 54.2 & \textbf{59.9} \bluegain{10.5} \\
      \texttt{causal\_judgment} & causal and commonsense &  65.3 & \textbf{68.4} \bluegain{4.7 }\\
        \texttt{english\_proverbs} & commonsense, contextual QA &	\textbf{91.2} & 87.5 \purplelost{4.2} \\
    \texttt{geometric\_shapes}    & algorithmic, visual &44.0 & \textbf{49.3} \bluegain{12.0} \\
    \texttt{physics\_questions} & logical, physics, math &7.6 & \textbf{12.5} \bluegain{64.5} \\
     \texttt{snarks} &commmonsense &69.1   &\textbf{86.1} \bluegain{24.6} \\
          \texttt{analogical\_similarity} & analogical &36.5 & \textbf{37.5} \bluegain{2.7} \\
          \texttt{international\_phonetic\_alphabet\_nli} &reading comprehension & 65.9 & \textbf{68.0} \bluegain{3.2} \\
          \texttt{movie\_dialog\_same\_or\_different} & commonsense, reading compre. &  64.8 & \textbf{68.8} \bluegain{6.2} \\
              \texttt{timedial} & commonsense, logical &78.3& \textbf{81.2} \bluegain{3.7}\\
                  \texttt{question\_selection} & reading comprehension& 54.8& \textbf{59.8} \bluegain{9.1} \\
                  \texttt{logical\_fallacy\_detection} &   logical reasoning &80.3 & \textbf{81.4} \bluegain{1.4} \\
                  \texttt{unit\_interpretation} &  arithmetic, logical & 47.0 & \textbf{51.0} \bluegain{8.5} \\ 
                     \texttt{language\_identification} & multilingual & 36.0 & \textbf{38.9} \bluegain{8.1} \\
                     \midrule
                     average (21 tasks) & - & 64.3 & \textbf{67.7} \bluegain{5.3} \\
        \bottomrule
    \end{tabular}
    \caption{List of challenging tasks in the BigBench emergent suite (BBES) and corresponding scores of PaLM 540B and \modelname 540B. All results are reported with standard 5-shot prompting.} 
    \label{tab:bbtable}
\end{table}

We select a suite of challenging tasks from BigBench based on a criterion that performance on PaLM on these tasks remain relatively flat-lined at 8B and 62B scale but suddenly unlocks at 540B.  We also consider tasks that are difficult for PaLM 540B to solve (near random performance). We call these suite of tasks \textsc{emergent} suite of BigBench tasks (BBES) as inspired by the criterion set by \citet{wei2022emergent}. Note that while these set of tasks overlap but are not entirely identical to BBH~\citep{bbcot}. Morever, BBES uses the default prompting and templates as BIG-Bench and do not use chain-of-thought prompting. Hence, they are not entirely comparable. BBH results can be found later in section \ref{cotsection}.

\subsubsection{BIG-Bench results}
\begin{figure}[t]
     \centering
     \includegraphics[width=\textwidth]{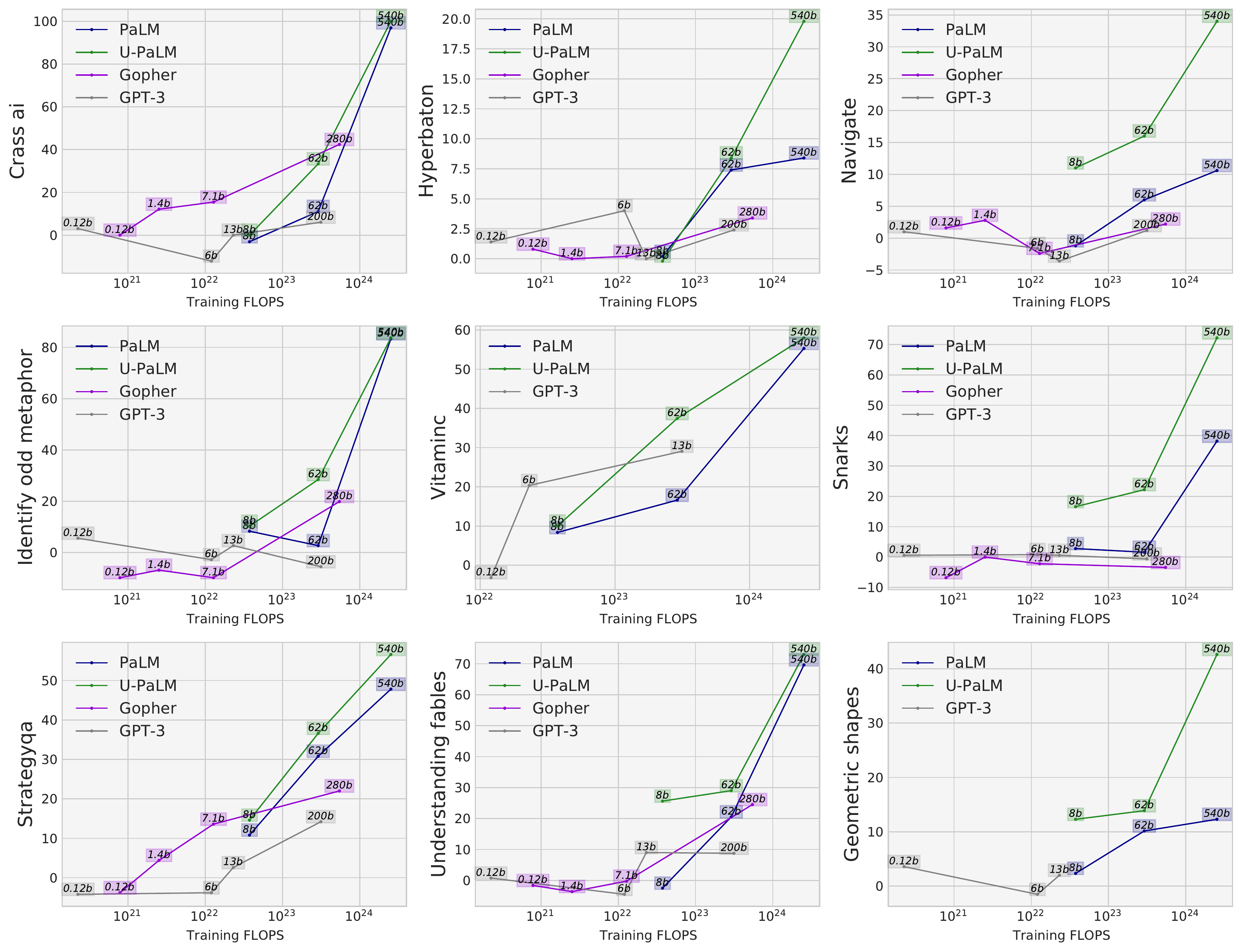}
    \caption{Scaling plots on BIG-Bench emergent suite (BBES) for different sizes of PaLM, \modelname, Gopher, and GPT-3 as a function of training FLOPs. Scores are normalized scores where zero denotes more or less random performance. X-axis is in log-scale.}
    \label{fig:downstream_scaling}
\end{figure}

Table \ref{tab:bbtable} reports the results of PaLM 540B and \modelname 540B on the BigBench emergent suite. We also describe the task and reasoning task for each task. Note that some tasks require a conjunction of various \textit{`skills'} to excel at. For example, the \texttt{navigate} task is a combination of spatial reasoning and arithmetic (counting). 
\paragraph{Overall results and Scaling Plots}
We observe that \modelname outperforms PaLM on 19 out of the 21 tasks at 540B scale. Moreover, the gains on certain tasks are substantial (e.g., $55.3\% \rightarrow 67.0\%$) on \texttt{navigate} and $69.1\% \rightarrow 86.1\%$ on \texttt{snarks}). On average, there is a $+5.4\%$ relative quality gain on the un-normalized aggregated average across all 21 tasks which we consider to be pretty strong results. Figure \ref{fig:downstream_scaling} which shows the scaling plots of \modelname relative to other models. Whenever possible, we also include baselines such as GPT-3 or Gopher from the official BIG-Bench repository. 
\paragraph{\methodname unlocks emergent task performance at smaller scales} Scale (e.g., scaling to 540B) is known to be one factor that results in emergent task performance~\citep{wei2022emergent}. We show that \methodname is able to elicit emergent abilities at smaller scales. For example, the quality on certain tasks such as \texttt{crass\_ai}, \texttt{vitaminc}, \texttt{identify\_odd\_metaphors} are tasks where performance starts to spike at 62B scale (as opposed to only at 540B with the PaLM model. In rarer occasions, the performance of \modelname 8B is even higher than PaLM 62B (e.g., \texttt{snarks}, \texttt{understanding\_fables}). Overall, these results show that there are strong evidence that \textit{inductive bias} (e.g., combinations of prefix language modeling, span corruption based pretraining in UL2) could be crucial when it comes to unraveling new abilities in large language models.

\subsubsection{Analyzing individual task performance on BIG-Bench}
This section dives into individual task performance and attempts to understand quality on different types of BIG-Bench tasks.

\textbf{Spatial or Visual Reasoning Tasks} The first category of tasks that \modelname does extremely well on are tasks that require some form of spatial or visual reasoning (e.g., \texttt{navigate} or \texttt{geometric\_shapes}). In both of these tasks, \modelname 8B outperforms PaLM 540B. We postulate that this is due to the prefix language model architecture and additional PrefixLM training that \modelname undergoes. To give a better illustration, consider the following examples from these tasks.
\begin{itemize}
\item In the \texttt{navigate} task, an example is as follows: \textit{`Turn right. Take 1 step. Turn right. Take 6 steps. Turn right. Take 1 step. Turn right. Take 2 steps. Take 4 steps.`} and the task is a binary classification task that determines if the agent returns to the starting point.
\item In the \texttt{geometric\_shapes} task, the goal is to predict the shape given an SVG path, e.g., given \textit{`M 31,29 L 34,76 L 82,16 L 31,29'} the model should predict \textit{triangle}.
\end{itemize}
Here, it is worth noting that both tasks can be improved intuitively by having bidirectional attention and being trained using a PrefixLM like objective. This could explain why \modelname could outperform PaLM 540B even at 8B because it was given the right inductive bias.

\textbf{Commonsense and Knowledge Tasks}
A reasonable portion out of the 21 tasks require some form of commonsense or language-based knowledge in order to do well. It is worth noting that \modelname does not train on any new unique tokens (or new data) and therefore, has no access to no new \textit{`knowledge'} compared to vanilla PaLM. Hence, gains here are expected to be milder compared to tasks that rely more on algorithmic or other types of reasoning. However, we observe some relatively smaller gains in certain tasks (e.g., \texttt{understanding\_fables} or \texttt{movie\_dialog\_same\_or\_different}). Amongst the tasks in this category, one exception is the \texttt{snarks} task which involves detecting sarcasm in natural language. It is worth noting that the only 2 out of 21 tasks where \modelname underperforms PaLM belongs to this category (e.g., \texttt{logical\_sequence} and \texttt{english\_proverbs}). We think this is reasonable since we do not completely expect \methodname to \textit{always} improve upon this category of tasks given that it does not actually process new data tokens.

\textbf{Context Reasoning or Reading Comprehension Tasks} Some tasks require some understanding of context and then requires the language model to answer questions based on this context. An example of this is the \texttt{vitaminc\_fact\_verficiation} task which tries to determine the veracity of a claim given external evidence (context). Another example is the \texttt{understanding\_fables} task where the goal is to determine the \textit{`morale of the story'} given context (passage or story). It is worth noting that \modelname exhibits emergence at 62B scale on these two tasks even though the final 540B model performance is relatively similar. We postulate that this is due to the architectural (and pretraining) advantage of PrefixLM which aids the model in performing much better even at smaller scales. Intuitively, being able to bidirectionally reason with context (prefix) could be important in context reasoning tasks.

\textbf{Multi-step Reasoning, Analogical Reasoning and Arithmetic tasks} We observe that there are some performance improvements on analogical reasoning task (e.g., \texttt{analogical\_similarity}) or multi-step reasoning tasks (\texttt{strategyqa}) at 540B scale. However, unlike context reasoning tasks, the performance on these class of tasks tend to follow similar scaling patterns albeit with slightly better performance. For example, based on Figure \ref{fig:downstream_scaling}, we note that \texttt{strategyqa} follows relatively similar scaling curves to PaLM.

\subsection{Zero-shot and Few-shot NLP}
In this section, we evaluate our models on various well-established NLP tasks. These tasks test a spectrum of zero and few-shot abilities of \modelname.
\subsubsection{Commonsense Reasoning}
We conduct experiments on four zero-shot commonsense reasoning benchmarks. Specifically, following~\citep{hoffmann2022training}, we use BoolQ~\citep{clark2019boolq}, PIQA~\citep{bisk2020piqa}, HellaSWAG~\citep{zellers2019hellaswag} and Winogrande~\citep{sakaguchi2019winogrande}. Aside from PaLM 62B and PaLM 540B which we use for direct comparisons with \modelname, we also compare with Chinchilla 70B~\citep{hoffmann2022training} and Gopher 280B~\citep{rae2021scaling}.  Table \ref{tab:commonsense} reports the results on zero-shot commonsense reasoning.

\begin{table}[H]
    \centering
    \small
    \begin{tabular}{l|cccccc}
    \toprule
        Task / Model & PaLM& \modelname & Chinchilla & Gopher & PaLM & \modelname \\
        Size &  62B& 62B& 70B & 280B & 540B & 540B \\
           FLOPS (ZFLOPS) & 295.7 & 298.7 & 588 & 504& 2527.2 & 2529.7 \\
        \midrule
       BoolQ 0-shot  &84.8  &85.4 & 83.7 & 81.8 & 88.0 & \textbf{88.8} \bluegain{0.9} \\
       PIQA 0-shot & 80.5& 81.4 &81.8 & 81.8 & 82.3 & \textbf{84.1} \bluegain{2.2} \\
       HellaSwag 0-shot & 79.7& 79.7 & 80.8&79.7&83.4 & \textbf{84.1} \bluegain{0.8} \\ 
       Winogrande 0-shot & 77.0 & 76.2&74.9 & 70.1&81.1 & \textbf{82.6} \bluegain{1.8} \\ 
       \midrule
Avg. Commonsense & 80.5 & 80.7 & 80.3 & 78.2 &83.7 & \textbf{84.9} \bluegain{1.4} \\       
       \bottomrule
    \end{tabular}
    \caption{Results on zero-shot commonsense reasoning.}
    \label{tab:commonsense}
\end{table}
 We show that \modelname 540B outperforms PaLM 540B on all four tasks with an average of (+1.4\%) relative improvement and attains the best performance across all models.

\subsubsection{Question Answering and Reading Comprehension}
We evaluate zero-shot and few-shot closed book question answering (CBQA) tasks~\citep{nqa, joshi2017triviaqa, roberts2020much} along with the zero-shot Lambada reading comprehension task~\citep{paperno-etal-2016-lambada}. Table \ref{tab:cbqa} reports the results of our experiments. We compare with PaLM 62B, PaLM 540B, Chinchilla 70B and Gopher 280B. 
\begin{table}[H]
    \centering
    \small
    \begin{tabular}{l|cccccc}
    \toprule
        Task / Model & PaLM& \modelname & Chinchilla & Gopher & PaLM & \modelname \\
        Size &  62B& 62B& 70B & 280B & 540B & 540B \\
        FLOPS (ZFLOPS) & 295.7 & 298.7 & 588 & 504& 2527.2 & 2529.7 \\
        \midrule
TriviaQA 0-shot & 67.3 & 68.3 & 67.0 & 52.8 & \textbf{76.9} & 76.4 \purplelost{0.7} \\ 
TriviaQA few-shot & 72.7& 73.6& 73.2 & 63.6&81.4 & \textbf{82.0} \bluegain{0.7} \\
Natural Questions 0-shot & 18.1 & 18.7 & 16.6 &10.1 &21.2 & \textbf{21.7} \bluegain{2.4} \\ 
Natural Questions few-shot & 27.6 & 30.5 & 31.5 &24.5 &36.0 & \textbf{40.1} \bluegain{11.4} \\
Lambada 0-shot & 75.4 & 79.7 & 77.2 & 74.5 & 77.9 & \textbf{80.5} \bluegain{3.3} \\
\midrule
Avg. QA/RC & 52.2 & 54.3 & 53.0 & 45.1&58.7 & \textbf{60.1} \bluegain{2.3}\\
       \bottomrule
    \end{tabular}
    \caption{Results on closed book QA and reading comprehension.}
    \label{tab:cbqa}
\end{table}
Overall, on few-shot CBQA and reading comprehension, we observe that \modelname 540B outperforms PaLM 540B by +2.3\% on average and up to $+11.4\%$ on few-shot natural questions. Meanwhile, the gain at 62B scale is also strong (i.e., +$2.1\%$ on average).

\subsubsection{Reasoning and Chain-of-thought Experiments}
\label{cotsection}
We conduct experiments on reasoning and CoT and compare \modelname 540B with PaLM 540B and Minerva 540B. We use the GSM8K~\citep{cobbe2021training}, BBH~\citep{bbcot}, StrategyQA \citep{geva2021did} and CommonsenseQA \citep{talmor-etal-2019-commonsenseqa} benchmarks. All tasks are run with chain-of-thought (CoT) prompting.
\begin{table}[H]
    \centering
    \small
    \begin{tabular}{l|ccc}
    \toprule
      Task / Model  & Minerva 540B &  PaLM 540B & \modelname 540B \\
      \midrule
        GSM8K& 57.8 & 54.9 &\textbf{58.5} \bluegain{6.6} \\
        BBH & 37.2 &44.8 & \textbf{49.6} \bluegain{10.7} \\
        StrategyQA & 61.9 &  76.4 & \textbf{76.6} \bluegain{0.2} \\
        CSQA & 72.2 & 76.9 & \textbf{80.1} \bluegain{4.2} \\
        \bottomrule
    \end{tabular}
    \caption{Experiment results on reasoning and chain-of-thought reasoning experiments.}
    \label{tab:reasoning_exps}
\end{table}
Table \ref{tab:reasoning_exps} reports results on reasoning and CoT benchmarks. \modelname 540B outperforms both PaLM 540B and Minverva 540B. Notably, the gains on GSM8K and BBH are relatively strong. This shows that \modelname does well on reasoning and is well-suited for chain-of-thought reasoning.

\subsubsection{Multilingual Few-shot Reasoning and Question Answering Tasks}
We conduct experiments on few-shot multilingual reasoning and question answering tasks. We use the MGSM (multilingual grade school math) benchmark proposed in~\citep{shi2022language}. For multilingual question answering, we use the well-established TydiQA~\citep{clark2020tydi} benchmark. In our experiments, both PaLM 540B and \modelname 540B uses chain-of-thought prompting~\citep{wei2022chain}. 
\begin{table}[H]
    \centering
    \small
    \begin{tabular}{c|cc}
    \toprule
  Task /  Model     & PaLM 540B & \modelname 540B \\
    \midrule
      TydiQA   & 52.9 & \textbf{54.6} \bluegain{3.2} \\
      MGSM & 45.9 & \textbf{49.9} \bluegain{8.7} \\

      \bottomrule
    \end{tabular}
    \caption{Experiments on Multilingual GSM (MGSM)~\citep{shi2022language} and TydiQA~\citep{clark2020tydi}}
    \label{tab:multilingual_tasks}
\end{table}
Table \ref{tab:multilingual_tasks} reports our results on MGSM and TydiQA. Our results show that \modelname outperform PaLM by a considerable margin (+3.2\% on TydiQA and +8.7\% on MGSM).

\subsubsection{Massively Multi-Task Language Understanding}
We compare PaLM and \modelname on the Massively Multi-Task Language Understanding (MMLU) benchmark~\citep{hendrycks2020measuring}. Table \ref{tab:mmlu} reports our results on MMLU's test set. Prior results are reported from~\citep{hoffmann2022training}. Our results show that \modelname outperforms PaLM on this task in the 5-shot setup by $2.0\%$ relative gain.
\begin{table}[H]
    \centering
    \begin{tabular}{l|c}
    \toprule
    Method & Accuracy \\
    \midrule
      Random  & 25.0\%  \\
    Average Human Rater & 34.5\%\\
    GPT-3 5-shot & 43.9\% \\ 
    Gopher 5-shot & 60.0\% \\ 
    Chinchilla 5-shot & 67.6\% \\
    \midrule
    PaLM 540B 5shot &  69.3 \% \\
    \modelname 540B 5-shot & \textbf{70.7} \% \bluegain{2.0}\\ 
    \bottomrule
    \end{tabular}
    \caption{Results on Massively Multi-Task Language Understanding (MMLU) test set.}
    \label{tab:mmlu}
\end{table}
\subsection{Finetuning}
We conduct experiments on SuperGLUE~\citep{wang2019superglue} and TydiQA~\citep{clark2020tydi} finetuning. We conduct experiments at 8B and 62B scale\footnote{Finetuning at 540B is compute intensive and probably less relevant for the finetuning setup since large scale LMs are typically used for prompting. Meanwhile, it is significantly more likely that smaller models are fine-tuned.}. Fine-tuning is conducted with a constant learning rate for $100k$ steps with a batch size of $32$.
\begin{table}[H]
    \centering
    \small
    \begin{tabular}{l|cccc}
    \toprule
         &  PaLM 8B & \modelname 8B & PaLM 62B & \modelname 62B \\
         \midrule
     SuperGLUE (Avg)   & 83.4 & \textbf{86.1} \bluegain{3.2} & 89.5 & \textbf{91.4} \bluegain{2.1} \\
     TydiQA (EM/F1) & 75.7 / 85.2 & \textbf{77.5} \bluegain{2.3} / \textbf{86.7} \bluegain{1.7} & 78.3 / 87.3 & \textbf{78.4} \bluegain{0.1} / \textbf{88.5} \bluegain{2.1} \\ 
         \bottomrule
    \end{tabular}
    \caption{Results on finetuning on SuperGLUE and TydiQA dev sets.}
    \label{tab:finetuning}
\end{table}
Table \ref{tab:finetuning} reports finetuning results. We observe that there is substantial improvement in fine-tuning especially at the 8B scale. The gains diminish slightly at 62B scale but are still modest in general. We note that PaLM's fine-tuning performance can be generally considered weaker than expected. For instance, PaLM 8B is generally outperformed by a T5 large model on the SuperGLUE dev average. We postulate that training PaLM on UL2 and span corruption tasks in complement to causal language modeling can ameliorate some of its flaws. Our results ascertains this by showing that \modelname{} strongly improves quality especially at smaller (8B) scales.

\section{Qualitative Analysis: New Prompting Capabilities}

Beyond improving the scaling behavior of PaLM, we find that the small amount of continued training applied in \methodname is sufficient to imbue PaLM with new prompting abilities introduced by the UL2 objective. Namely, the use of denoising in UL2 allows PaLM to acquire infilling abilities. Infilling allows \modelname to have a second approach to tackling prompts, which we observe to be very useful. In addition, with \modelname we can also supply mode tokens to gain access to specific pretraining objectives. This gives us a powerful tool to control the model without making any updates to the model or its inference. In this section we provide some examples of situations where \modelname's expanded prompting capabilities prove to be useful. 

\subsection{Infilling Ability}
Left-to-right casual language model pretraining has typically allowed models to provide meaningful continuations of prompts. With \modelname we observe that, by extending pretraining with a small amount of UL2 denoising steps, the model is also able to pick up infilling abilities -- where the model is given a location in the middle of a prompt to fill in. Notably, with \modelname it is possible to query both the infill style and the traditional style via the usage of extra ID tokens (as it is used in denoising) or without, respectively.

\begin{figure}[H]
\includegraphics[width=\textwidth]{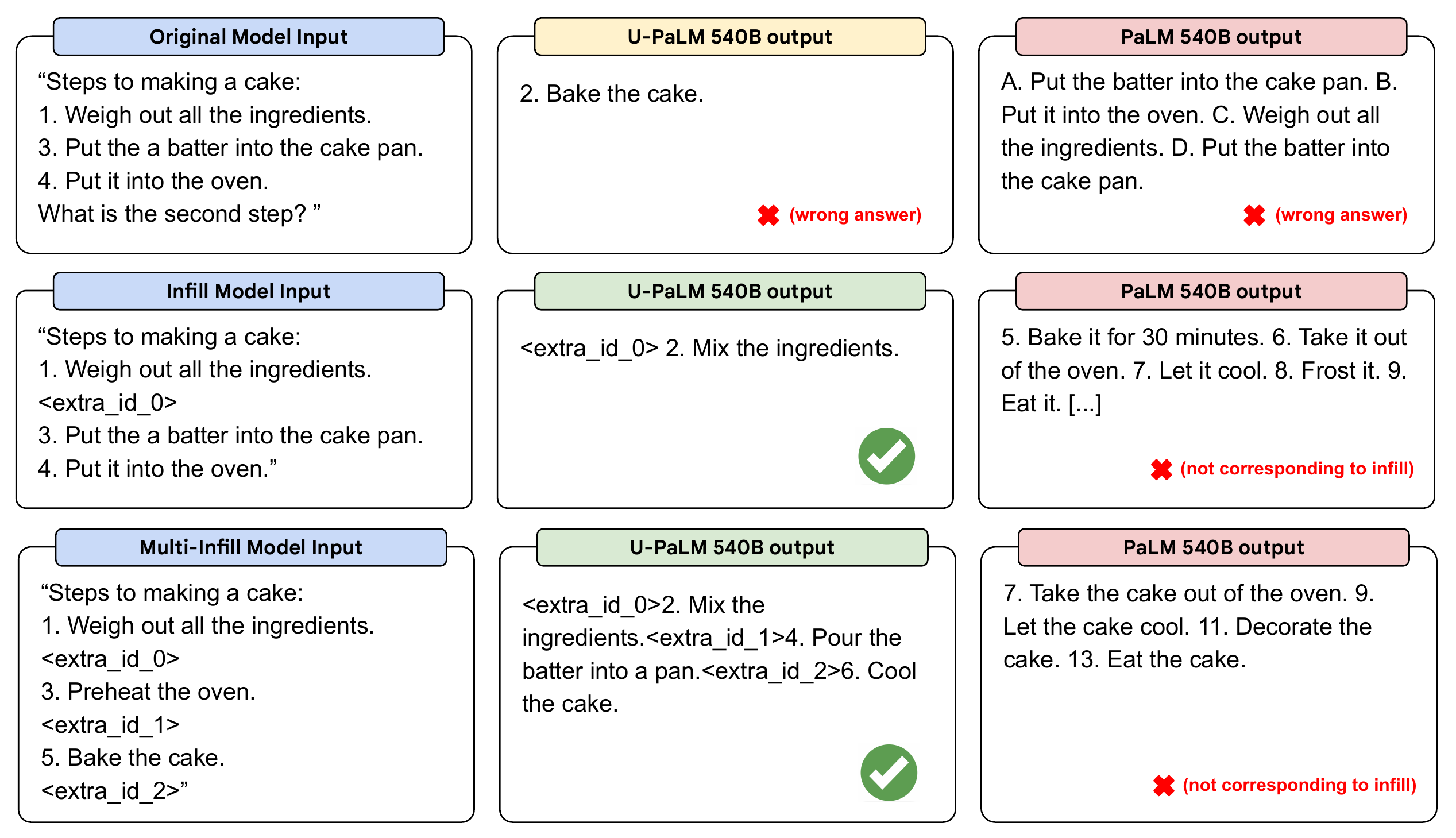}
\caption{An example of a prompt that is improved by rephrasing to use \modelname's infilling capabilities.}
\label{fig:infill-examples}
\end{figure}

In Figure \ref{fig:infill-examples}, we include example outputs for PaLM, \modelname with traditional prompting, as well as \modelname with infill prompting. We phrase this particular prompt in two ways: one as a question that is suitable for traditional prompting via PaLM and one leveraging \modelname's infill capabilities. In the traditional phrasing, both PaLM and \modelname do not produce the correct answer. With the infill phrasing, PaLM ignores the infill token (extra ID token) as PaLM has not seen it during training, and instead produces the rest of the steps after step 4. \modelname correctly infills the second step in this example. Finally, a third example is included to demonstrate \modelname's ability to infill multiple slots. These examples demonstrate that, with only a small amount of additional training, we are able to expand the functionality of PaLM to serve an entirely new class of queries.

\subsection{Leveraging Specific Pretraining Modes}
Recall that via the UL2 objective, R-, X-, and S-denoisers are associated with the [NLU], [NLG], and [S2S] mode tokens respectively. S-denoisers are essentially the PrefixLM objective, while R- and X-denoisers are variations of span corruption, and thus are also associated with extra ID tokens which we can use during prompting for infill (as shown above.) Given this unique setup, we can control the mode token during inference to gain access to specific knowledge that might have been acquired in one mode but not another. This effectively provides us with more options in how to answer prompts, without the need to make any changes to the learned model or its inference algorithm.
\begin{figure}[H]
\includegraphics[width=\textwidth]{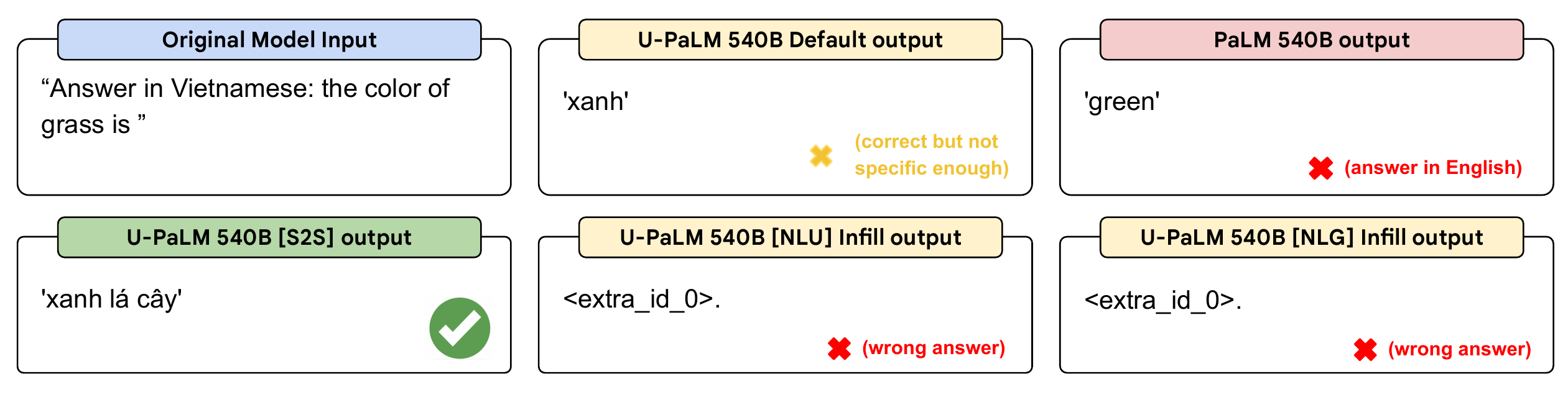}
\caption{An example of a prompt that works only when querying a specific pretraining mode.}
\label{fig:mode-infill-examples}
\end{figure}

In Figure \ref{fig:mode-infill-examples}, we include a challenging example where we ask the model to do zero-shot cross-lingual question answering from an English question into a Vietnamese answer. For PaLM and \modelname default, we pass the input as-is to the model. For the rest, we prepend one of [S2S], [NLU], or [NLG] to the beginning of the input, and in the case of [NLU] and [NLG], we add the infill token at the end of the input, as typical for these modes. Interestingly, \modelname in [S2S] mode is the only variant that returns the correct answer in Vietnamese. Regular PaLM produces the correct answer, but ignores the Vietnamese request, while \modelname with default prompting (no mode, no infill) produces a roughly correct answer but could be more specific ('xanh' encompasses both greens and blues). This example shows how accessing specific mode tokens may work well for some prompts more so than others, giving us a powerful technique to serve a larger variety of prompts.

\subsection{Improved Diversity for Open-ended Generation}
\begin{figure}[H]
\includegraphics[width=\textwidth]{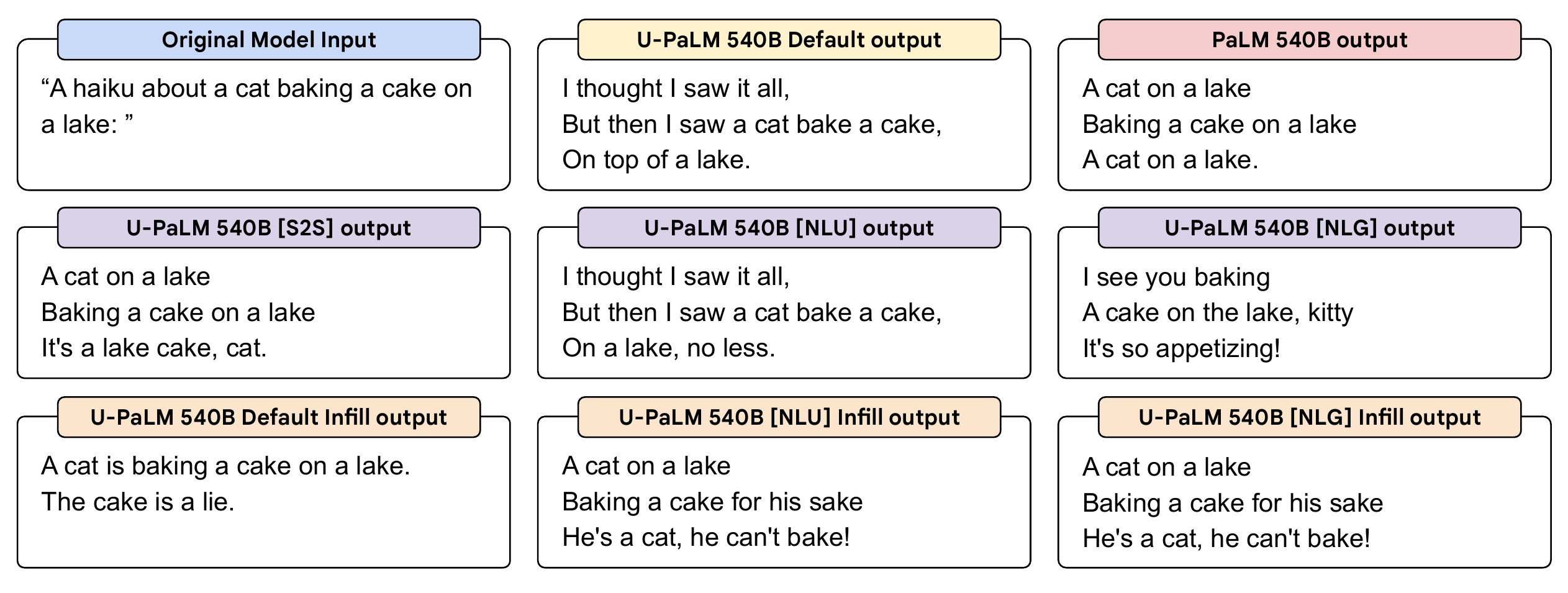}
\caption{An example of querying \modelname for multiple diverse outputs by using different prompt mode token and LM/infill combinations.}
\label{fig:haiku-examples}
\end{figure}

Even though [NLU] and [NLG] modes typically coincide during pretraining with span corruption (involving extra ID tokens, infilling), we can still use [NLU] and [NLG] mode tokens with no infilling at all. Similarly we can use infilling but with no mode tokens. The variety of ways to prompt \modelname results in a useful technique to increase the diversity of the outputs we can get from the model, without resorting to alternative decoding techniques (e.g. sampling). This is particularly useful for more open-ended prompts.

In Figure \ref{fig:haiku-examples}, we ask PaLM and all variants of querying \modelname to write a haiku about "a cat baking a cake on a lake" - a very random prompt that the model is unlikely to see during training, yet requires very structured output. All outputs use greedy decoding here, and surprisingly all models generate reasonable haikus about the topic, although not all follow a strict 5-7-5 syllable structure. PaLM's haiku repeats the first and last line, which is somewhat less interesting. We can see that the different combinations of querying \modelname results in pleasantly varying poems.

\section{Conclusion and Future Work}
We proposed \methodname for continued training of PaLM. We show that with only $\approx0.1\%$ additional FLOPs (or compute), we are able to improve the scaling curve and properties of PaLM on many downstream tasks and metrics. Notably, \methodname{} enables a 4.4 million TPUv4 savings at 540B scale. The resulting model which we call \modelname outperforms PaLM on English NLP tasks (e.g., commonsense reasoning and closed-book question answering), reasoning tasks with chain-of-thought, multilingual reasoning, MMLU and a suite of challenging BIG-Bench tasks. 

Also, does \methodname play well with multi-task fine-tuning methods like FLAN? Normally, we would like to end with a cliche future work statement here but not today because we have done this already here \citep{chung2022flan}. So check it out!

\section*{Acknowledgements}
We thank Le Hou and Oliver Bousquet for their advice and feedback on the paper. We thank Barret Zoph and William Fedus for early discussions about this paper.  We thank Adam Roberts for feedback on prior work.

\bibliographystyle{plainnat}
\bibliography{ref}
\newpage
\section{Appendix}

\subsection{Details of Scaling Curves for Few-shot Experiments}
\label{app:fewshot-exp-details}
We compute a mean aggregated score of the following tasks. We use 21 zero-shot rank classification tasks, i.e., BoolQ, Record, COPA, RTE, WiC, WSC, CB, MultiRC, Winograd, Winogrande, ANLI R1, ANLI R2, ANLI R3, PIQA, StoryCloze, HellaSwag, Arc-E, Arc-C, RaceM, RaceH, OpenbookQA. We use 5 one-shot generative tasks, i.e., TriviaQA, NaturalQuestions, WebQuestions,SQuaDV2 and Lambada. All tasks use the accuracy (or exact match) metric except MultiRC which reports f1a following~\citep{brown2020language}. In total, the aggregated metric is a mean over all \textbf{26} tasks. We list the scores that correspond to Figure \ref{fig:scaling}'s 540B scaling plot below.

\begin{table}[H]
    \centering
    \begin{tabular}{l|ccc|ccc}
    \toprule
    Model & \multicolumn{3}{c}{PaLM 540B}  & \multicolumn{3}{c}{\modelname 540B}\\ 
        Task/\#Tokens & 182B  & 329B & 780B  & 182B$^+$ & 329B$^+$ & 780B$^+$   \\
        \midrule
         TriviaQA 1shot & 73.4 & 74.4 & 81.4 & 73.3 & 75.6 & \textbf{82.0} \\
         NQA 1shot & 23.2 & 25.6 & 29.3 & 24.4 & 28.1 & \textbf{30.7}\\
         WebQA 1shot & 21.6 & 19.9 & 22.6 & 21.0 & 21.7 & \textbf{23.4}\\ 
         BoolQ & 82.4	&  85.6 & 88.0 & 85.8 & 88.2 & \textbf{88.8}\\
         ReCORD & 91.5 &  92.7 & 92.9 & 91.5 & 92.6 & \textbf{93.0}\\
         COPA & 92.0 &  93.0 & 93.0 & 94.0 & 93.0 & \textbf{96.0}\\
         RTE & 68.6 & 67.2 & 72.9 & 73.7 & 71.5 & \textbf{75.5}\\
         WIC & 50.8 & 53.8 & 59.1 & 52.2 & 58.0 & \textbf{62.2}\\
         WSC & 88.1 & 86.7 &\textbf{89.1} & 87.0 & 88.1 & 87.4\\
         CB &	57.1 & 48.2 & 51.8 & 69.6 & 71.4 & \textbf{69.6}\\
         MultiRC & 	76.7 & 81.1 & 83.5 & 78.4 & 81.7 & \textbf{83.8}\\
         Winogrande & 89.4 & 88.3 & \textbf{90.1} & 87.9 & 89.7 & 88.3\\
         Winograd & 76.9 & 79.6 & 81.1 & 78.2 &  79.3 & \textbf{82.6}\\
         ANLI R1 & 44.3 &49.4 & 48.4 & 50.3 & 50.6 & \textbf{55.3}
         \\
         ANLI R2 & 41.3 & 42.7 & 44.2 & 43.5 & 45.2 & \textbf{47.8}\\
         ANLI R3 & 43.8 & 42.8 & 45.7 & 46.7 & 49.3 & \textbf{57.0}\\ 
         PIQA & 81.0 & 81.9 & 82.3 & 80.8  & 82.0 & \textbf{84.1}\\
         StoryCloze & 82.7  & 83.9 & 84.6 & 83.7 & 84.2 & \textbf{87.0}\\
         HellaSwag & 79.1 & 81.8 & 83.4 & 79.5 & 82.3 & \textbf{84.1}\\ 
         ArcE & 74.8 & 72.8 & 76.6 & 74.6 & 76.3 & \textbf{85.9}\\
         ArcC &48.0 & 46.9 & 53.0 & 48.6 & 50.4 & \textbf{60.3}\\
         RaceM & 63.6 & 67.3 & \textbf{68.1} & 63.2 & 67.1 & 67.2\\
         OpenbookQA & 50.2 & 51.2 & 53.4 & 50.2 & 51.2 & \textbf{53.6}\\
         RaceH & 45.3 & 48.5 & 49.1 & 45.5 & 48.5 & \textbf{51.3}\\
         Lambada 1shot & 75.4 & 77.5 & \textbf{81.8} & 74.3 & 79.9 &80.0 \\
         SquadV2 1shot & 70.5 & 71.3 & \textbf{78.7} & 71.8 & 70.3 & 78.2\\
         \midrule 
         Average & 62.7 & 63.8 & 66.5 & 64.1 &  66.2 & 69.4 \\
         \bottomrule
    \end{tabular}
    \caption{Results of PaLM vs \modelname at different FLOPs (\# tokens) at 540B scale.}
    \label{tab:my_label}
\end{table}

\subsection{Details of Vocab and Sentinel Tokens}
For \modelname, we had to train on span corruption or infilling task. We use the same setup as UL2 and T5 where we inject sentinel tokens, e.g., \textit{<extra\_id\_0>} into the masked positions. In T5, sentinel ids are added as 100 additional vocab tokens at the end of the sentencepiece (vocab). In PaLM, since we restart from an existing PaLM checkpoints, it was quite cumbersome to initialize 100 new embeddings in the vocab. Hence, we opt to simply use the last 100 subwords as sentinel tokens. Finally, we also use eos symbols in the vocab when training the model.

\subsection{Additional Discussion}
In this section, we delve into some additional topics and discussions.
\subsubsection{What about training from scratch?}
We address the elephant in the room. There are multiple perspectives to this question. The first is that \methodname can be thought as a form of \textit{`UL2 schedule`} that sets a single causal language model objective from 0 to $N$ steps and then doing the UL2 mixture from $N$ to $N+\epsilon$. In this sense, if we wanted to train from scratch, this would require modifying the mixture to have significantly more causal language modeling. The second perspective is that \methodname introduces a natural curriculum where the model spents a large fraction of training acquiring basic language modeling before moving on to tasks like infilling or learning how to leverage bidirectional receptive fields. Whether there is a taxonomy or hierarchical of pretraining tasks is still an open question which we hope to answer in future work. The third perspective is simply the practical aspect of \modelname. Training a PaLM 540B model from scratch is incredibly costly and we would like to reuse our existing models (or components) as much as possible to design new models for new tasks. \modelname is an instance of this type of research. Finally, given that many language models are trained as causal language models, we believe that \methodname presents great opportunity for improving existing models with only a small amount of compute.

\subsubsection{What about supervised finetuning on many new tasks like FLAN or T0?}
Glad you asked. Check out our other work \citep{chung2022flan} that shows \methodname and FLAN is complementary!

\end{document}